\documentclass[10pt,twocolumn,letterpaper]{article}

\usepackage[pagenumbers]{cvpr} 

\usepackage{times}
\usepackage{epsfig}
\usepackage{graphicx}
\usepackage{amsmath}
\usepackage{amssymb}
\usepackage{paralist}
\usepackage{float}
\usepackage{stfloats}
\usepackage{booktabs}
\usepackage{multicol}
\usepackage{multirow}
\usepackage{bm}
\usepackage{bbm}
\usepackage{diagbox}
\usepackage{makecell}
\usepackage{algorithm}  
\usepackage{algorithmic} 
\usepackage{color, colortbl}
\usepackage{marvosym}
\newcommand{\gr}{\rowcolor[gray]{.95}} 
\newcommand{\good}{\rowcolor[RGB]{234,243,233}}
\definecolor{cvprblue}{rgb}{0.21,0.49,0.74}
\usepackage[pagebackref,breaklinks,colorlinks,allcolors=cvprblue]{hyperref}

\def\ours{R4Det}
\newcommand{\figref}[1]{Figure~\ref{#1}}%
\newcommand{\tabref}[1]{Table~\ref{#1}}%

\renewcommand{\eqref}[1]{Eq.~(\ref{#1})}

\def\paperID{19822} 
\def\confName{CVPR}
\def\confYear{2026}

\title{R4Det: 4D Radar-Camera Fusion for High-Performance 3D Object Detection}

\author{Zhongyu Xia$^1$\quad Yousen Tang$^1$\textsuperscript{†}\quad Yongtao Wang$^1$\textsuperscript{\Letter}\quad Zhifeng Wang$^2$\quad Weijun Qin$^2$
\\
$^1$ Wangxuan Institute of Computer Technology, Peking University
\quad 
$^2$ EBTech Co. Ltd
\\
{\tt\small \{xiazhongyu,wyt\}@pku.edu.cn
\quad \tt\small tangyousen@tongji.edu.cn}
\quad \tt\small zhifengw@ebtech.com
}

\begin{document}
\maketitle

\renewcommand{\thefootnote}{}
\footnotetext{\textsuperscript{†}This work was done as an intern at PKU. ~\textsuperscript{\Letter}Corresponding author.}

\begin{abstract}

4D radar–camera sensing configuration has gained increasing importance in autonomous driving.
However, existing 3D object detection methods that fuse 4D Radar and camera data confront several challenges.
First, their absolute depth estimation module is not robust and accurate enough, leading to inaccurate 3D localization.
Second, the performance of their temporal fusion module will degrade dramatically or even fail when the ego vehicle's pose is missing or inaccurate. 
Third, for some small objects, the sparse radar point clouds may completely fail to reflect from their surfaces. In such cases, detection must rely solely on visual unimodal priors.
To address these limitations, we propose \ours, which enhances depth estimation quality via the Panoramic Depth Fusion module, enabling mutual reinforcement between absolute and relative depth. 
For temporal fusion, we design a Deformable Gated Temporal Fusion module that does not rely on the ego vehicle's pose.
In addition, we built an Instance-Guided Dynamic Refinement module that extracts semantic prototypes from 2D instance guidance. 
Experiments show that \ours~achieves state-of-the-art 3D object detection results on the TJ4DRadSet and VoD datasets.
The source code and models will be released at \url{https://github.com/VDIGPKU/R4Det}.

\end{abstract}

    
\section{Introduction}
\label{sec:intro}

The cornerstone of autonomous driving safety lies in 3D perception. 
To achieve this goal, the community has been actively exploring sensor configurations that go beyond the conventional LiDAR-camera setup. 
Among emerging alternatives, 4D millimeter-wave radar has gained significant attention for its all-weather reliability, long-range sensing, and substantially lower cost compared with LiDAR. 
However, the inherent sparsity and noise characteristics of radar point clouds make it insufficient to support high-precision 3D detection on its own. 
Therefore, 4D radar-camera multi-modal fusion approaches have become the prevailing choice in practical applications. 
However, developing high-performance 4D radar-camera 3D object detection models faces numerous challenges.

\begin{figure}[t]
    \centering
    \includegraphics[width=0.83\linewidth]{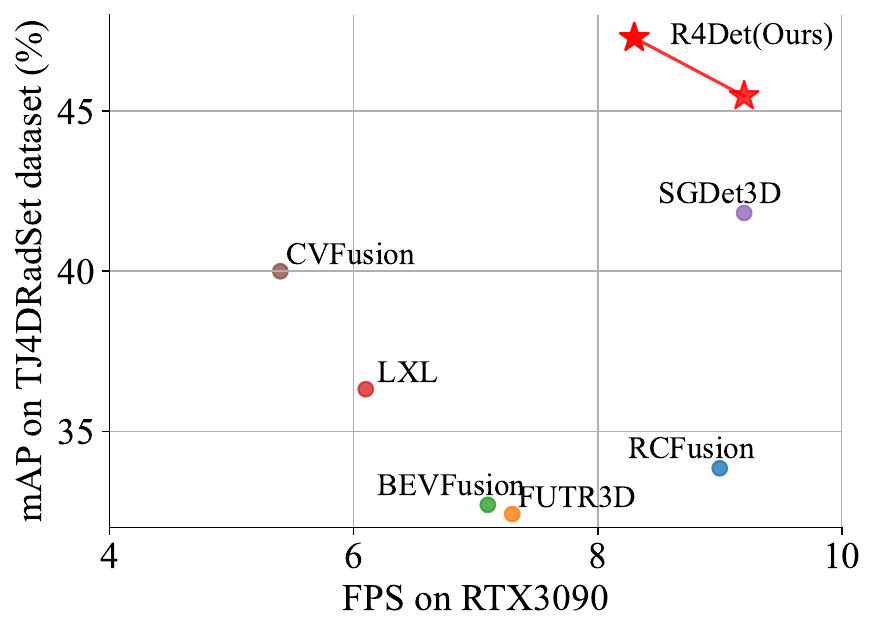}
    \vspace{-10pt}
    \caption{\textbf{Comparison of \ours~with current 4D radar-camera real-time detectors.}
    }
    \label{fig:multitask_result}
    \vspace{-15pt}
\end{figure}

The primary challenge in current 4D radar–camera fusion frameworks is the limited accuracy of their absolute depth estimation module, which leads to inaccurate 3D localization.
Current radar–camera fusion frameworks~\cite{RCBEVDet, bevfusion, sgdet3d} rely on accurate dense depth estimation during view transformation. 
However, these methods only apply absolute depth supervision to foreground points, resulting in sparse depth supervision. 
%
Hence, they can not achieve high-quality panoramic depth estimation, which benefits foreground localization and reduces background noise.
%
In addition, recent relative depth estimation works show strong generalization capabilities~\cite{bhat2023zoedepth, yang2024depth, yin2023metric3d}. How to effectively leverage their capabilities for accurate panoramic absolute depth estimation is a topic worth investigating.

The second challenge confronting current radar-camera fusion frameworks involves temporal fusion due to the absence of the ego-vehicle pose. 
Temporal input provides crucial historical 3D information of occluded objects, which is essential for 3D object detection tasks. 
However, most methods developed upon the mainstream 4D radar dataset, \textit{i.e.}, TJ4DRadSet, remain single-frame detectors, largely due to the lack of ego-vehicle pose in this dataset. 
Moreover, in real-world scenarios, particularly in rural areas, missing or inaccurate ego-vehicle poses due to GPS signal loss are also common. 
Hence, some studies~\cite{CRN} have begun incorporating temporal information by merely stacking historical BEV features along the channel dimension, which yield limited improvements.

The third challenge lies in detecting small objects, \textit{e.g.}, distant cyclists, which appear in the image without any point cloud reflected from their surfaces. 
For these challenging cases, it is necessary to leverage visual priors to identify them. 
Some works~\cite{jiang2024far3d, bevformerv2} attempt to extract instance proposals from 2D features, but they all rely on Transformer-based architectures and regard the extracted instances only as queries, making them less compatible with other widely used CNN-based frameworks.

To address the three challenges mentioned above, we propose \ours, a novel 4D radar-camera fusion framework for high-performance 3d object detection. 
To handle the first challenge, we propose the Panoramic Depth Fusion (PDF) module.
Specifically, to eliminate geometric contamination, PDF abandons sparse metric regression and introduces a novel pairwise ordinal ranking loss with dynamic tolerance, enabling the model to learn high-fidelity, edge-sharp, and structurally continuous geometry.
To tackle the second challenge, we design a Deformable Gated Temporal Fusion (DGTF) module that explicitly aligns non-rigid motion with deformable convolutions and smooths temporal noise through GRU-style gating, achieving stable feature aggregation without the instability of implicit recurrence.
To deal with the third challenge, we propose an Instance-Guided Dynamic Refinement (IGDR) that leverages a 2D instance-guided semantic prototype to purify features, jointly optimized with the detection head to correct misprojected semantics and restore clarity for distant or partially occluded objects.
To sum up, the key contributions of this work are:

\begin{compactitem}
    \item
    We present a 4D radar-camera detector for highly accurate, efficient, and robust 3D object detection. 

    \item
    We propose three BEV-paradigm-compatible, plug-and-play modules, \textit{i.e.}, Panoramic Depth Fusion, Deformable Gated Temporal Fusion, and Instance-Guided Dynamic Refinement, to produce high-quality panoramic absolute depth estimations, fuse temporal information without ego pose, and handle small objects, respectively.

    \item
    \ours~achieves state-of-the-art performance on both VoD and TJ4DRadSet dataset. 

    
\end{compactitem}

\section{Related Work}
\label{sec:related}
\subsection{Camera-only 3D Object Detection}
Current camera-only 3D detection methods focus on reconstructing the 3D scene from synchronized image streams. BEVDet~\cite{BEVDet} and BEVDepth~\cite{BEVDepth} predict dense pixel-wise depth maps and employ the Lift-Splat-Shoot (LSS) paradigm~\cite{LSS} to project image features into the bird's-eye-view (BEV) space, constructing dense BEV representations. Transformer-based methods, such as DETR3D~\cite{DETR3D} and PETR~\cite{PETR}, use BEV queries or object-centric queries to perform implicit 2D-to-3D coordinate transformations and feature aggregation. Sparse geometry sampling strategies have also been explored to balance feature efficiency and spatial precision. To capture temporal information in dynamic scenarios, BEVFormer~\cite{BEVFormer} and VideoBEV~\cite{VideoBEV} integrate temporal information either in the BEV space or via recurrent feature fusion. Despite their rich semantic representations, camera-only systems still suffer from significant depth and geometric uncertainty under occlusion, illumination changes, or distant objects. 

\subsection{4D Radar-Camera Fusion for 3D Object Detection}

To overcome the limitations of camera-only perception and integrate robust geometric and motion information, 4D millimeter-wave radar-camera fusion has attracted significant attention. 
BEV-level fusion methods, such as CRN~\cite{CRN}, use multi-modal deformable attention to align radar and camera BEV features, while HVDetFusion~\cite{HVDetFusion} and RCBEVDet~\cite{RCBEVDet} explore effective radar BEV feature extraction to supplement camera BEV streams. 
HyDRa~\cite{HyDRa} fuses features in both perspective and BEV spaces, CRAFT~\cite{CRAFT} introduces proposal-level fusion.
RADIANT~\cite{RADIANT} leverages radar points to predict 3D offsets for correcting monocular depth errors without retraining the camera model. 
SGDet3D~\cite{sgdet3d} introduces a dual-branch fusion module and a localization-aware cross-attention to facilitate deep interactions across modalities. 
CVFusion~\cite{cvfusion} proposed Point- and grid-guided fusion modules to effectively fuse multiple-view features.

However, existing fusion detectors still face three challenges: (1) depth estimation heavily relies on sparse radar or LiDAR supervision, which induces geometric contamination with blurred boundaries and discontinuous structures; (2) temporal modeling is often simplified as naive feature concatenation, failing to align non-rigid motion effectively; (3) cumulative errors in geometric and modality alignment can contaminate BEV features, impairing instance separation and the detection of distant small objects.


\begin{figure*}[!th]
    \centering
    \includegraphics[width=0.9\linewidth]{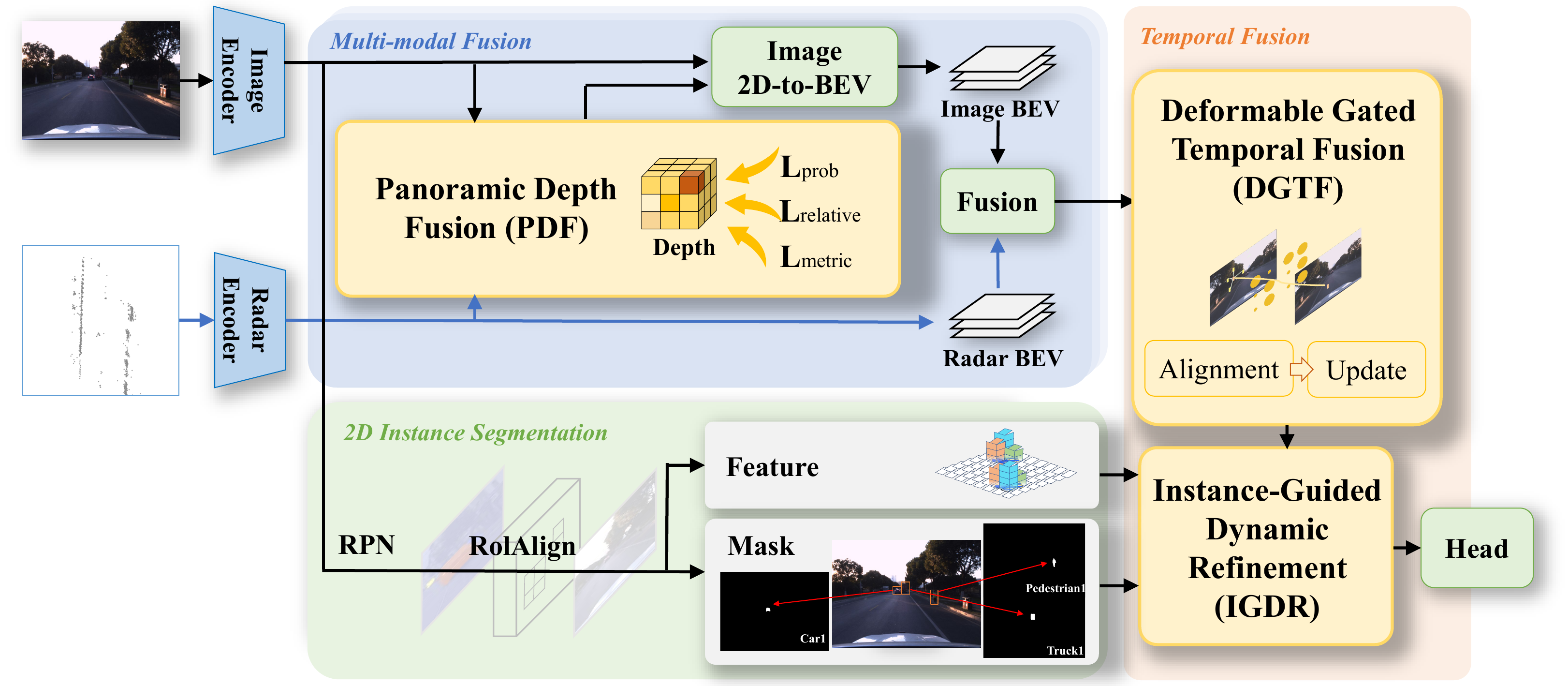}
    \vspace{-8pt}
    \caption{\textbf{Overall architecture of \ours.}  Our framework progressively purifies the BEV representation in three stages: i) The Panoramic Depth Fusion (PDF) module generates a geometrically-accurate BEV feature map from multi-modal inputs. 
    ii) The Deformable Gated Temporal Fusion (DGTF) module performs pose-free alignment and integration to create a temporally consistent feature. 
    iii) The Instance-Guided Dynamic Refinement (IGDR) module leverages 2D instance prototypes to purify the final features for 3D detection.}
    \label{fig:overall}
    \vspace{-15pt}
\end{figure*}

\section{Method}
\label{sec:method}

\subsection{Overall Framework}
As illustrated in \figref{fig:overall}, the \ours~framework is a progressive purification pipeline designed to address three core challenges in 4D radar-camera object detection. 
First, using sparse radar features (extracted by the radar encoder) as queries, the \textbf{Panoramic Depth Fusion (PDF)} module establishes an accurate geometric foundation to extract dense image semantics (Sec.~\ref{sec:32}). The resulting image BEV is then concatenated with the radar BEV via a \textit{Multi-modal Fusion} block to produce an initial fused BEV feature map $X_t$. 
Next, our \textbf{Deformable Gated Temporal Fusion (DGTF)} module performs pose-free spatial alignment and gated updates on $X_t$ to maintain a temporally consistent hidden state $H_t$ and output a fused feature $F_{RC}$ (Sec.~\ref{sec:33}). 
Finally, the \textbf{Instance-Guided Dynamic Refinement (IGDR)} module leverages 2D instance prototypes to dynamically calibrate $F_{RC}$, outputting a purified feature $F_{final}$ to the 3D detection head (Sec.~\ref{sec:34}).

%
\begin{figure}[t]
    \centering
    \includegraphics[width=\columnwidth]{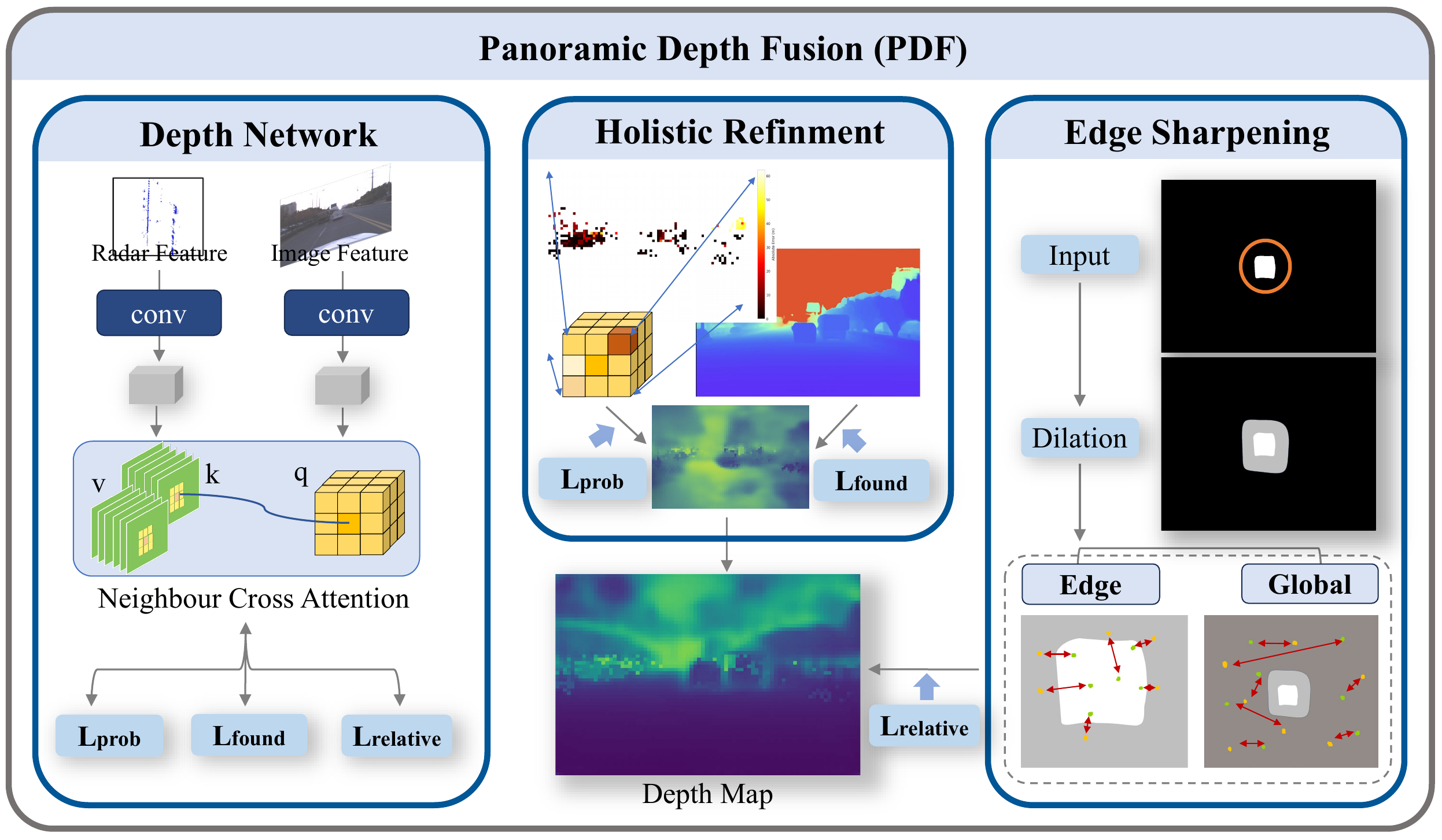}
    \vspace{-16pt}
    \caption{\textbf{Overview of  the Panoramic Depth Fusion (PDF) module.}}
    \label{fig:depth}
    \vspace{-15pt}
\end{figure}

\subsection{Panoramic Depth Fusion (PDF) }
\label{sec:32}

To establish a robust geometric foundation, we employ a Neighborhood Cross-Attention mechanism~\cite{sgdet3d} to aggregate dense image semantics using sparse radar features as queries. Built upon this aggregated representation, we introduce a triplet of dedicated supervisions to fundamentally resolve depth ambiguity and geometric contamination.
\noindent\textbf{Probabilistic Supervision.}
We first supervise the predicted depth distribution $\mathcal{P} \in \mathbb{R}^{B \times D \times H \times W}$ using sparse high-precision LiDAR points $d_g^{\text{sparse}}$. Here, $B$ is the batch size, $D$ represents the number of depth bins, and $H, W$ are the spatial dimensions of the BEV map.
Specifically, each sparse point is converted to a Gaussian target distribution $\mathcal{G}(d_g^{\text{sparse}})$.
Then, we minimize the KL divergence over sparse locations $\mathcal{M}_{\text{sparse}}$:
\begin{equation}
\mathcal{L}_{prob} = \frac{1}{|\mathcal{M}_{\text{sparse}}|} 
\sum_{i \in \mathcal{M}_{\text{sparse}}} 
\text{KL} \big( \mathcal{G}(d_{g_i}^{\text{sparse}}) \parallel \mathcal{P}_i \big).
\end{equation}
This branch determines the network's probabilistic output, ensuring that the `Splat' operation in the subsequent view transformation is based on a sharp and accurate distribution, thereby preventing geometric scattering.

\noindent\textbf{Foundation-Model-Guided Depth Supervision}
For the decoded expected depth map $\hat{d}$, we incorporate both sparse radar-anchored depth and dense pseudo-ground truth (Metric3D), denoted as $d_g^{\text{dense}}$, as supervision. By applying the Smooth L1 loss for both metric targets, we effectively prevent gradient explosion caused by potential radar outliers:
\begin{equation}
\mathcal{L}_{\text{found}} = \lambda_{\text{abs}} \mathcal{L}_{\text{abs}} + \lambda_{\text{dense}} \mathcal{L}_{\text{dense}},
\end{equation}
where
\begin{align}
\mathcal{L}_{\text{abs}} &= \frac{1}{|\mathcal{M}_{\text{sparse}}|} 
\sum_{i \in \mathcal{M}_{\text{sparse}}} 
\text{SmoothL1}(\hat{d}_i, d_{g_{i}}^{\text{sparse}}), \\
\mathcal{L}_{\text{dense}} &= \frac{1}{|\mathcal{M}_{\text{dense}}|} 
\sum_{i \in \mathcal{M}_{\text{dense}}} 
\text{SmoothL1}(\hat{d}_i, d_{g_{i}}^{\text{dense}}).
\end{align}
This combination ensures both high metric precision on key points and robust structural coverage for the full scene.

\noindent\textbf{Structural Ranking Supervision.}
To preserve local depth relationships, we adopt a relative depth ranking loss:
\begin{equation}
s_{ij} = \text{sign}(d_{g_i}^{\text{dense}} - d_{g_j}^{\text{dense}}),\quad
\end{equation}
\begin{equation}
\mathcal{L}_{pair}(i,j) = \text{Softplus}(- s_{ij} (\hat{d}_i - \hat{d}_j)).
\end{equation}
However, naively sampling all pairs is inefficient, as most pairs lie on flat surfaces where small true depth differences are dominated by sensor noise, leading to ambiguous training signals.
Therefore, we only select pixel pairs with a significant difference that exceeds their depth-dependent dynamic threshold given by:
\begin{equation}
\begin{split}
\tau_{ij} = \max(\tau_{abs}, \tau_{rel} \cdot (d_{g_i}^{\text{dense}} + d_{g_j}^{\text{dense}})/2), \\
\text{include pair if } |d_{g_i}^{\text{dense}} - d_{g_j}^{\text{dense}}| > \tau_{ij}.
\end{split}
\end{equation}
where $\tau_{abs}$ and $\tau_{rel}$ are the absolute and relative tolerance margins, respectively, designed to filter out noisy minor differences on flat surfaces.
Furthermore, we apply a \textit{foreground-biased dual sampling} strategy which separates edge regions and background:
\begin{equation}
\mathcal{L}_{relative} = w_{edge} \mathcal{L}_{edge} + w_{global} \mathcal{L}_{global}.
\end{equation}
Here, $w_{edge}$ and $w_{global}$ are the balancing weights. $\mathcal{L}_{global}$ samples pixel pairs randomly from background regions, excluding foreground objects, to preserve overall scene structure. 
The core innovation lies in $\mathcal{L}_{edge}$'s \textit{cross-boundary sampling}: we first dilate 2D instance masks to form a narrow \textit{dilated ring} around each object (the difference between the dilated mask and original mask). During sampling, one pixel $i$ is selected from this dilated ring (immediately outside the object), and the other $j$ from the object's interior (original mask). This strategy forces the network to focus on sharp depth transitions at object boundaries, significantly enhancing contour sharpness and geometric accuracy.

\noindent\textbf{Total Depth Loss.}
The overall depth loss combines all three supervisions is:
\begin{equation}
\mathcal{L}_{depth} = \lambda_1 \mathcal{L}_{prob} + \lambda_2 \mathcal{L}_{found} + \lambda_3 \mathcal{L}_{relative}.
\end{equation}
This depth loss encourages the network to generate depth maps that are probabilistically accurate, metric-scale precise, and structurally coherent, providing strong geometric priors for downstream 3D detection.

\begin{figure}[t]
    \centering
    \includegraphics[width=0.90\linewidth]{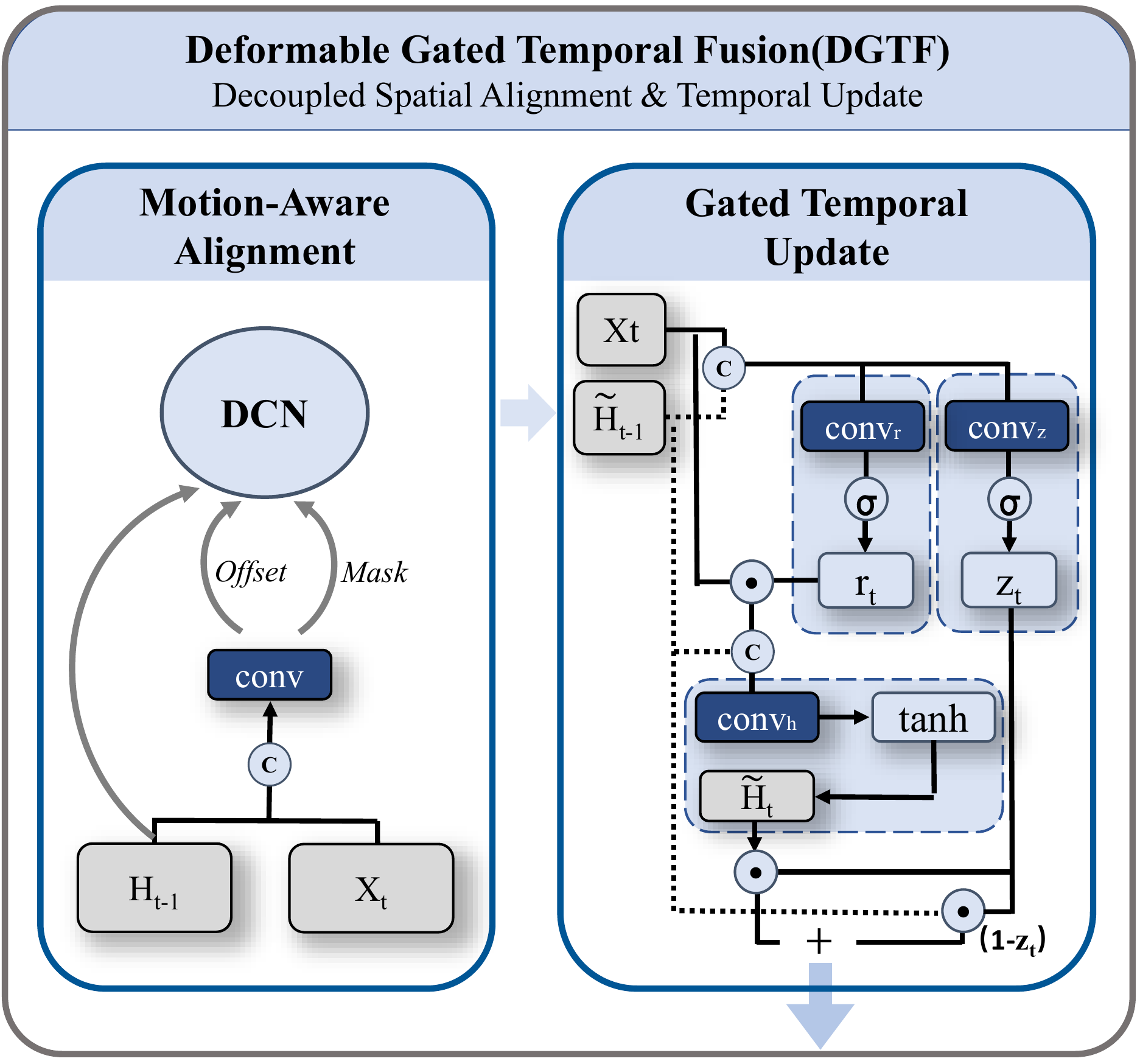}
    \vspace{-8pt}
    \caption{\textbf{Architecture of the proposed Deformable Gated Temporal Fusion (DGTF) module.} 
    DGTF consists of two specialized branches: motion-aware alignment using deformable convolution and a gated temporal update mechanism.}
    \label{fig:dgtf}
    \vspace{-10pt}
\end{figure}

\begin{algorithm}[t]
\caption{Pseudo-code of Deformable Gated Temporal Fusion (DGTF)
}
\label{alg:dgtf}
\begin{algorithmic}[1]
\footnotesize
\STATE \textbf{Input:} Current BEV feature $X_t$, previous hidden state $H_{t-1}$
\STATE \textbf{Output:} Updated BEV feature $H_t$
\STATE // \textit{Motion-Aware Alignment}
\STATE $(\Delta p, m) = \text{Conv}_{\text{offset}}(\text{Concat}(X_t, H_{t-1}))$
\STATE $\tilde{H}_{t-1} = \text{DCNv2}(H_{t-1}, \Delta p, m)$
\STATE // \textit{Gated Information Update}
\STATE $r_t = \sigma(\text{Conv}_r(\text{Concat}(X_t, \tilde{H}_{t-1})))$
\STATE $\tilde{H}_t = \phi(\text{Conv}_h(\text{Concat}(X_t, r_t \odot \tilde{H}_{t-1})))$
\STATE $z_t = \sigma(\text{Conv}_z(\text{Concat}(X_t, \tilde{H}_{t-1})))$
\STATE $H_t = (1 - z_t) \odot X_t + z_t \odot \tilde{H}_t$
\STATE $F_{RC} = \text{Conv}_{out}(H_t)$
\STATE \textbf{return} $F_{RC}$
\end{algorithmic}
\end{algorithm}

\subsection{Deformable Gated Temporal Fusion (DGTF)}
\label{sec:33}

Fusing dense BEV features over time introduces two coupled challenges: 
\textit{spatial misalignment} caused by ego-motion and object movement, and \textit{temporal inconsistency} due to dynamic scene evolution. 
Conventional recurrent units try to address both issues jointly through a single gating mechanism, 
which is often inefficient and inaccurate when handling complex non-rigid motion. 
To address this issue, we explicitly decouple the challenge into two problems and propose the Deformable Gated Temporal Fusion (DGTF) module with two branches, as illustrated in~\figref{fig:dgtf}. 
The motion-aware alignment branch addresses the spatial correction issue, while the gated temporal update branch addresses the feature evolution problem.

\noindent\textbf{Motion-Aware Alignment Branch.}
Given the current BEV feature $X_t$ and the previous hidden state $H_{t-1}$, 
we first predict the sampling offset $\Delta p$ and modulation mask $m$ through an offset prediction convolution:
\begin{equation}
\Delta p, m = \text{Conv}_{offset}(\text{Concat}(X_t, H_{t-1})).
\end{equation}
The previous hidden feature is then aligned using deformable convolution (DCNv2):
\begin{equation}
\tilde{H}_{t-1} = \text{DCNv2}(H_{t-1}, \Delta p, m).
\end{equation}
This step compensates for motion-induced spatial misalignment without ego-pose priors: the learned offsets $\Delta p$ explicitly reconstruct the relative motion flow, while the modulation mask $m$ adaptively suppresses unreliable background regions, thereby producing geometrically consistent features across frames.

\noindent\textbf{Gated Information Update Branch.}
After feature alignment, DGTF performs temporal reasoning via gated update units.
The reset gate $r_t$ estimates the relevance of the aligned historical feature:
\begin{equation}
r_t = \sigma(\text{Conv}_r(\text{Concat}(X_t, \tilde{H}_{t-1}))).
\end{equation}
The candidate feature $\tilde{H}_t$ integrates the filtered history and current observation:
\begin{equation}
\tilde{H}_t = \phi(\text{Conv}_h(\text{Concat}(X_t, r_t \odot \tilde{H}_{t-1}))).
\end{equation}
Finally, the update gate $z_t$ adaptively balances between new and historical information:
\begin{equation}
z_t = \sigma(\text{Conv}_z(\text{Concat}(X_t, \tilde{H}_{t-1}))),
\end{equation}
\begin{equation}
H_t = (1 - z_t) \odot X_t + z_t \odot \tilde{H}_t.
\end{equation}
where $\sigma$ and $\phi$ denote the sigmoid and hyperbolic tangent (tanh) activation functions, respectively, and $\odot$ represents the element-wise multiplication. The refined feature $H_t$ is further processed by a lightweight convolutional layer($\text{Conv}_{out}$) to produce the temporally aligned multi-modal feature $F_{RC}$ for the subsequent module.

By disentangling spatial alignment and temporal update through 
``DCN for alignment, GRU for update'', 
DGTF achieves fine-grained motion compensation and robust temporal consistency, 
leading to significantly improved dynamic perception in radar-camera BEV fusion.

\subsection{Instance-Guided Dynamic Refinement (IGDR)}
\label{sec:34}

\begin{figure*}[t]
    \centering
    \includegraphics[width=0.9\linewidth]{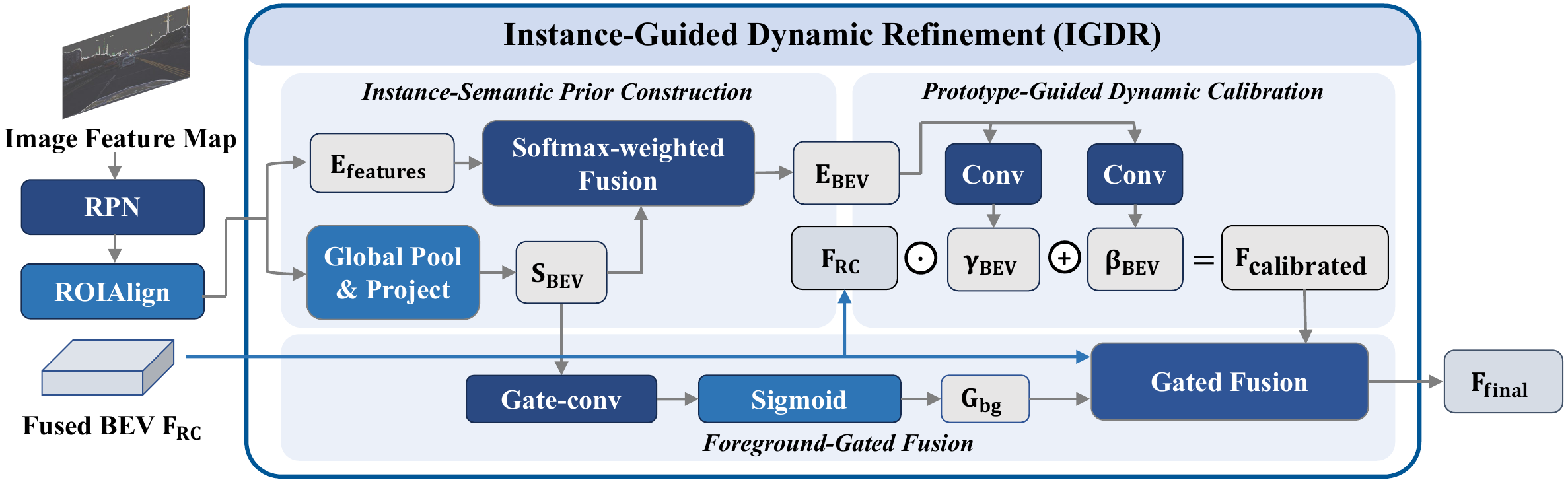}
    \vspace{-6pt}
    \caption{\textbf{Overview of Instance-Guided Dynamic Refinement (IGDR) module.} 
    IGDR adaptively refines radar-camera BEV features by suppressing instance overlap contamination and cross-modality noise, while preserving reliable distant object representations.}
    \label{fig:dcr}
    \vspace{-10pt}
\end{figure*}

To address feature contamination in overlapping regions and blurred representations of distant small objects in BEV, we propose the Instance-Guided Dynamic Refinement (IGDR) module. The core idea is guidance: instead of relying on potentially noisy $F_{RC}$ features to refine themselves, we leverage parallel, clean 2D instance features as semantic priors to actively calibrate the main BEV feature stream.

\noindent\textbf{Instance-Semantic Prior Construction.}
IGDR takes three inputs: the multi-modal BEV feature $F_{RC} \in \mathbb{R}^{B \times C \times H \times W}$ (from DGTF), 2D instance branch features $E_{features} \in \mathbb{R}^{N \times C_{inst} \times H' \times W'}$, and the instance spatial distribution map $S_{BEV} \in \mathbb{R}^{B \times N \times H \times W}$. Here, $C$ denotes the channel dimension of the BEV map,$N$ and $C_{inst}$ represent the number of RPN instance proposals and their channel dimension, while $H'$ and $W'$ are the spatial resolutions of the 2D RoI features extracted by the 2D instance segmentation head. $S_{BEV}$ serves as a spatial assignment map, representing the probability distribution of the $N$ instances projected into the BEV space via LSS.

We first apply global average pooling over the spatial dimensions of $E_{features}$ and a projection layer to align channel dimensions, yielding instance prototype vectors $E_{proj} \in \mathbb{R}^{B \times N \times C_{inst}}$. Using $S_{BEV}$ as a soft attention map, we broadcast the prototypes into the BEV space via \textit{Softmax-weighted Fusion} to construct a clean instance feature map $E_{BEV}$:
\begin{equation}
A_{prob} = \text{Softmax}_{dim=N}\left(\frac{S_{BEV}}{\tau}\right), \quad
\end{equation}
\begin{equation}
E_{BEV} = \text{BMM}(A_{prob}, E_{proj}).
\end{equation}

Here, $\tau$ is a temperature hyperparameter controlling the attention sharpness. $E_{BEV} \in \mathbb{R}^{B \times C_{inst} \times H \times W}$ represents the ideal instance semantics, fully decoupled from potential geometric projection errors and noise in $F_{RC}$.

\noindent\textbf{Prototype-Guided Dynamic Calibration.}
The core innovation of IGDR is instance-guided dynamic calibration. Rather than directly fusing $E_{BEV}$ with $F_{RC}$, we treat $E_{BEV}$ as a conditional generator to predict a pair of affine transformation parameters (scale $\gamma_{BEV}$ and bias $\beta_{BEV}$) for each spatial location via \textit{Conv} layers:
\begin{equation}
\gamma_{BEV} = \text{Conv}_{\gamma}(E_{BEV}), \quad
\beta_{BEV} = \text{Conv}_{\beta}(E_{BEV})
\end{equation}

We then apply a feature-wise affine transformation to the potentially noisy $F_{RC}$, producing the calibrated feature $F_{calibrated}$:

\begin{equation}
F_{calibrated} = F_{RC} \odot \gamma_{BEV} + \beta_{BEV}
\end{equation}

This mechanism allows the clean instance semantics to actively \textit{calibrate, enhance, or suppress} the main feature stream, effectively mitigating cross-instance contamination and improving the representation of distant small objects.

\noindent\textbf{Foreground-Gated Fusion.}
To ensure calibration only affects instance regions while preserving background structure, we introduce a foreground gate $G_{bg}$. We first sum $S_{BEV}$ along the instance dimension to represent the union of all instances, then apply a $3 \times 3$ Gate-conv layer followed by a \textit{Sigmoid} function ($\sigma$) to generate a smooth gating map:

\begin{equation}
G_{bg} = \sigma\left(\text{Conv}_{gate}\left(\sum_{i=1}^{N} S_{BEV}^{(i)}\right)\right)
\end{equation}

The final output is obtained via \textit{Gated Fusion}:

\begin{equation}
F_{final} = (1 - G_{bg}) \odot F_{RC} + G_{bg} \odot F_{calibrated}
\end{equation}
To prevent exposure bias during training, we strictly avoid using ground-truth bounding boxes to extract instance features. Instead, IGDR is trained using realistic, potentially imperfect proposals dynamically generated by the 2D detector. This simulated inference path forces the module to learn robust calibration under non-ideal priors, yielding a highly plug-and-play refinement block without modifying the backbone.
\section{Experiments}
\label{sec:exp}

\begin{table*}[!b]
\centering
\caption{\textbf{Comparison of 3D object detection results on the {\tt{test}} set of TJ4DRadSet \cite{tj4dradset}.}
In the modality column, R denotes 4D radar and C denotes camera. The best values are in \textbf{bold}.}
\label{tab:sota_tj4dradset}
\vspace{-10pt}
\resizebox{\textwidth}{!}{
\begin{tabular}{l|c|cccc|c|cccc|c}
\toprule
\multirow{2}{*}{\textbf{Method}} & \multirow{2}{*}{\textbf{Modality}} & \multicolumn{4}{c|}{\textbf{$\text{AP}_{\text{3D}}$ (\%)}} & \multirow{2}{*}{\textbf{$\text{mAP}_{\text{3D}}$}} & \multicolumn{4}{c|}{\textbf{$\text{AP}_{\text{BEV}}$ (\%)}} & \multirow{2}{*}{\textbf{$\text{mAP}_{\text{BEV}}$}} \\
\cline{3-6} \cline{8-11}
 & & Car & Pedestrian & Cyclist & Truck & & Car & Pedestrian & Cyclist & Truck & \\
\midrule

\gr PointPillars \cite{pointpillars} & R & 21.26 & 28.33 & 52.47 & 11.18 & 28.31 & 38.34 & 32.26 & 56.11 & 18.19 & 36.23 \\
CenterPoint \cite{centerpoint} & R & 22.03 & 25.02 & 53.32 & 15.92 & 29.07 & 33.03 & 27.87 & 58.74 & 25.09 & 36.18 \\
\gr RadarPillarNet \cite{rcfusion} & R & 28.45 & 26.24 & 51.57 & 15.20 & 30.37 & 45.72 & 29.19 & 56.89 & 25.17 & 39.24 \\
SMURF \cite{smurf} & R & 28.47 & 26.22 & 54.61 & 22.64 & 32.99 & 43.13 & 29.19 & 58.81 & 32.80 & 40.98 \\
\gr RCFusion \cite{rcfusion} & R+C & 29.72 & 27.17 & 54.93 & 23.56 & 33.85 & 40.89 & 30.95 & 58.30 & 28.92 & 39.76 \\
FUTR3D \cite{futr3d} & R+C & - & - & - & - & 32.42 & - & - & - & - & 37.51 \\
\gr BEVFusion \cite{bevfusion} & R+C & - & - & - & - & 32.71 & - & - & - & - & 41.12 \\
LXL \cite{lxl} & R+C & - & - & - & - & 36.32 & - & - & - & - & 41.20 \\
\gr SGDet3D \cite{sgdet3d} & R+C & 59.43 & 26.57 & 51.30 & 30.00 & 41.82 & 66.38 & 29.18 & 53.72 & 39.36 & 47.16 \\
CVFusion \cite{cvfusion} & R+C & 51.54 & 29.49 & 49.41 & 29.55 & 40.00 & 58.07 & 31.65 & 51.29 & 35.29 & 44.07 \\
\good \textbf{\ours} & \textbf{R+C} & \textbf{63.60} & \textbf{31.24} & \textbf{62.84} & \textbf{31.46} & \textbf{47.29} & \textbf{72.36} & \textbf{33.48} & \textbf{64.58} & \textbf{45.85} & \textbf{54.07} \\
\hline
\end{tabular}
}
\vspace{-7pt}
\end{table*}

\begin{table*}[t]
\centering
\caption[results_vod]{\textbf{Comparison of 3D object detection results on the {\tt{val}} set of VoD~\cite{vod}.}
R and C denote 4D radar and camera modalities, respectively. The best values are in \textbf{bold}.}
\vspace{-10pt}
\label{tab:sota_vod}
\resizebox{\textwidth}{!}{
\begin{tabular}{l|c|cccc|cccc|c}
\toprule
\multirow{2}{*}{\textbf{Methods}} & \multirow{2}{*}{\textbf{Modality}} &
\multicolumn{4}{c|}{\textbf{Entire Annotated Area ($\text{AP}_{\text{EAA}}$, \%)}} &
\multicolumn{4}{c|}{\textbf{Driving Corridor ($\text{AP}_{\text{DC}}$, \%)}} &
\multirow{2}{*}{\textbf{FPS}} \\
\cline{3-10}
 & & Car & Pedestrian & Cyclist & mAP & Car & Pedestrian & Cyclist & mAP & \\
\midrule

PointPillars~\cite{pointpillars} & R & 37.06 & 35.04 & 63.44 & 45.18 & 70.15 & 47.22 & 85.07 & 67.48 & 113.9 \\
\gr CenterPoint~\cite{centerpoint} & R & 32.74 & 38.00 & 65.51 & 45.42 & 62.01 & 48.18 & 84.98 & 65.06 & - \\
RadarPillarNet~\cite{rcfusion} & R & 39.30 & 35.10 & 63.63 & 46.01 & 71.65 & 42.80 & 83.14 & 65.86 & 98.8 \\
\gr SMURF~\cite{smurf} & R & 42.31 & 39.09 & 71.50 & 50.97 & 71.74 & 50.54 & 86.87 & 69.72 & - \\
RCFusion~\cite{rcfusion} & R+C & 41.70 & 38.95 & 68.31 & 49.65 & 71.87 & 47.50 & 88.33 & 69.23 & 9.0 \\
\gr FUTR3D~\cite{futr3d} & R+C & 46.01 & 35.11 & 65.98 & 49.03 & 78.66 & 43.10 & 86.19 & 69.32 & 7.3 \\
BEVFusion~\cite{bevfusion} & R+C & 37.85 & 40.96 & 68.95 & 49.25 & 70.21 & 45.86 & 89.48 & 68.52 & 7.1 \\
\gr LXL~\cite{lxl} & R+C & 42.33 & 49.48 & 77.12 & 56.31 & 72.18 & 58.30 & 88.31 & 72.93 & 6.1 \\
SGDet3D~\cite{sgdet3d} & R+C & 53.16 & 49.98 & 76.11 & 59.75 & 81.13 & 60.91 & 90.22 & 77.42 & 9.2 \\
\gr CVFusion~\cite{cvfusion} & R+C & 60.87 & \textbf{57.89} & 77.46 & 65.41 & 89.86 & \textbf{68.79} & 88.62 & 82.42 & 5.4 \\
\good \textbf{\ours} & R+C & \textbf{66.90} & 55.42 & \textbf{77.75} & \textbf{66.69} & \textbf{90.62} & 66.47 & \textbf{93.96} & \textbf{83.68} & 8.3 \\
\bottomrule
\end{tabular}
}
\vspace{-10pt}
\end{table*}
\begin{table}[t]
\centering
\caption{\textbf{More 3D object detection results on the {\tt{val}} set of VoD to demonstrate the proposed modules are plug-and-play.}}
\label{tab:onbevfusion}
\vspace{0pt}
\resizebox{\columnwidth}{!}{
    \begin{tabular}{l|ccc|l}
        \toprule
        Method & Car & Pedestrian & Cyclist & $\text{mAP}_{\text{EAA}}$  \\
        \midrule
        BEVFusion~\cite{bevfusion} & 37.85 & 40.96 & 68.95 & 49.25\\
        \good\textbf{BEVFusion + Ours} & 41.98 & 48.15 & 76.64 & \textbf{55.59}  \\
        \midrule
        RCBEVDet~\cite{RCBEVDet} & 40.63 & 38.86 & 70.48 & 49.99  \\
        \good\textbf{RCBEVDet + Ours} & 41.02 & 48.85 & 76.12 & \textbf{55.33}  \\
        \bottomrule
    \end{tabular}
}
\vspace{-12pt}
\end{table}

\subsection{Datasets and Metrics}
We conduct experiments on two large-scale 4D radar-camera datasets: TJ4DRadSet \cite{tj4dradset} and View-of-Delft \cite{vod} . TJ4DRadSet provides data from diverse driving conditions with four categories: \textit{Car}, \textit{Pedestrian}, \textit{Cyclist}, and \textit{Truck}. VoD offers data from complex urban traffic with three categories: \textit{Car}, \textit{Pedestrian}, and \textit{Cyclist}. We follow the official evaluation protocols for both. For TJ4DRadSet, we report $\mathrm{AP}_{\text{3D}}$ and $\mathrm{AP}_{\text{BEV}}$ for targets within 70m. For VoD, we report two complementary metrics: $\mathrm{AP}_{\text{EAA}}$ (Entire Annotated Area) and $\mathrm{AP}_{\text{DC}}$ (Driving Corridor). A unified IoU threshold convention is used: 0.50 for \textit{Car} and \textit{Truck}, and 0.25 for \textit{Pedestrian} and \textit{Cyclist}.

\subsection{Implementation Details}  

We train the model with a global batch size of 32 using a two-stage strategy identical to SGDet3D~\cite{sgdet3d}: (i) a 15-epoch spatial perception pretraining stage (freezing DGTF, IGDR, and the detection head) to initialize the PDF module and 2D instance branch for high-quality geometric and semantic priors; (ii) a 15-epoch spatial-temporal end-to-end finetuning stage with all parameters unfrozen to jointly optimize fusion, alignment, and dynamic refinement. We use AdamW with an initial learning rate of $4\times10^{-4}$ and cosine decay, along with standard data augmentations including random flipping, rotation, and scaling.
The data point on the right in Fig. 1 (mAP 45.47, FPS 9.2) reduced the channel numbers in its backbone and neck (256$\rightarrow$224) to align its FPS with SGDet3D.






\begin{figure*}[t]
    \centering
    \includegraphics[width=\linewidth]{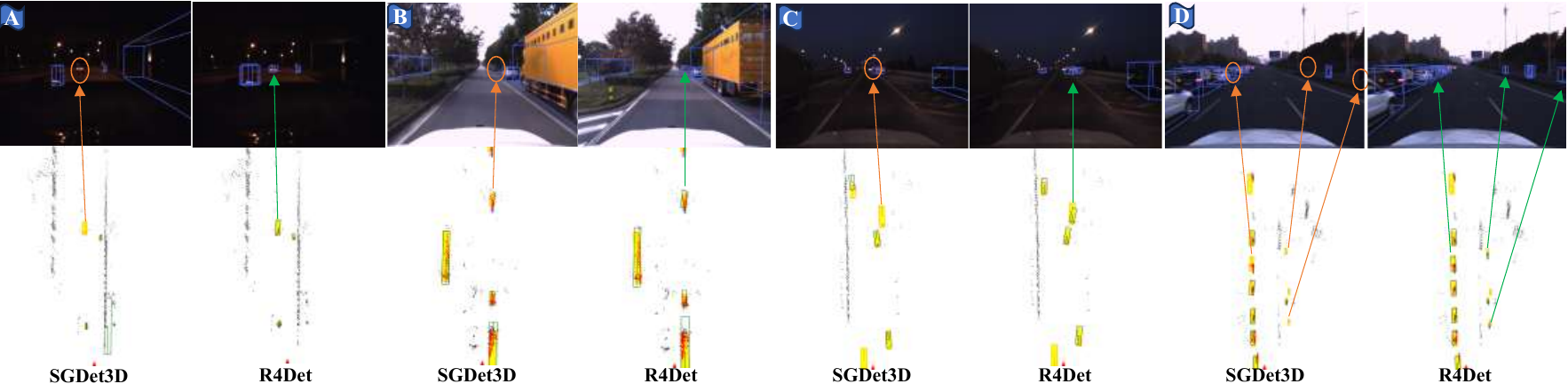}
    \vspace{-10pt}
    \caption{\textbf{Example visualization results of \ours~and baseline on challenging scenarios (\textit{e.g.}, low-light conditions or small objects).}
    }
    \label{fig:vis}
    \vspace{-12pt}
\end{figure*}

\subsection{Main Results}

As shown in~\tabref{tab:sota_tj4dradset}, \ours~significantly outperforms all existing methods across all categories on the challenging \textbf{TJ4DRadSet} dataset, which includes complex nighttime and overpass scenarios. 
Our framework improves upon the SGDet3D~\cite{sgdet3d} baseline by \textbf{+5.47\%} and \textbf{+6.91\%} in 3D and BEV mAP.
It is worth noting that our method is effective for small objects, such as \textit{Cyclist}, surpassing the previous best by a large margin of +7.91\%.
For the \textbf{VoD dateset}, as shown in~\tabref{tab:sota_vod}, \ours~also demonstrate state-of-the-art performance. 
As shown in~\figref{fig:multitask_result} and~\tabref{tab:sota_vod}, we conducted inference speed tests on a single 3090 GPU. The experimental results demonstrate that R4Det achieves a better speed-accuracy balance.

We also applied the proposed module to BEVFusion~\cite{bevfusion} and RCBEVDet~\cite{RCBEVDet}. As shown in~\tabref{tab:onbevfusion}, our approach achieves improvements of +6.34\% and +5.34\% in $\text{mAP}_{\text{EAA}}$, respectively. This demonstrates the plug-and-play nature of our method.



\subsection{Ablation}

We conduct a chained ablation study on the TJ4DRadSet validation set~\cite{tj4dradset}.
As shown in Table 1, the first row presents the SGDet3D baseline.
PDF enhances geometric perception through fine-grained depth completion (+1.71 mAP), DGTF strengthens temporal consistency via gated recurrent fusion (+3.55 mAP), and IGDR further refines BEV representations with instance-guided calibration (+3.66 mAP).

\begin{table}[t]
\centering
\caption{\textbf{Ablation study on each component of \ours~on the TJ4D Dataset}.}
\label{tab:overall_ablation}
\vspace{-10pt}
\setlength{\tabcolsep}{3pt}
\resizebox{\columnwidth}{!}{
\begin{tabular}{ccc|cc|cccc}
\toprule
\textbf{PDF} & \textbf{DGTF} & \textbf{IGDR} & \textbf{$\text{mAP}_{\text{BEV}}$} & \textbf{$\text{mAP}_{\text{3D}}$} & \textbf{Car} & \textbf{Ped} & \textbf{Cyc} & \textbf{Truck}\\
\midrule
 & & & 45.15 & 39.86 & 58.67 & 25.93 & 47.89 & 26.24\\
 \checkmark & & & 46.86 & 41.41 & 59.76 & 23.35 & 55.08 & 29.25\\
 \checkmark & \checkmark & & 50.41 & 44.86 & 61.94 & 28.90 & 57.15 & 30.48\\
 \checkmark & \checkmark & \checkmark & \textbf{54.07} & \textbf{47.29} & \textbf{63.60} & \textbf{31.24} & \textbf{62.84} & \textbf{31.46}\\
\bottomrule
\end{tabular}
}
\vspace{-10pt}
\end{table}

\begin{figure}[t]
    \centering
    \includegraphics[width=0.98\linewidth]{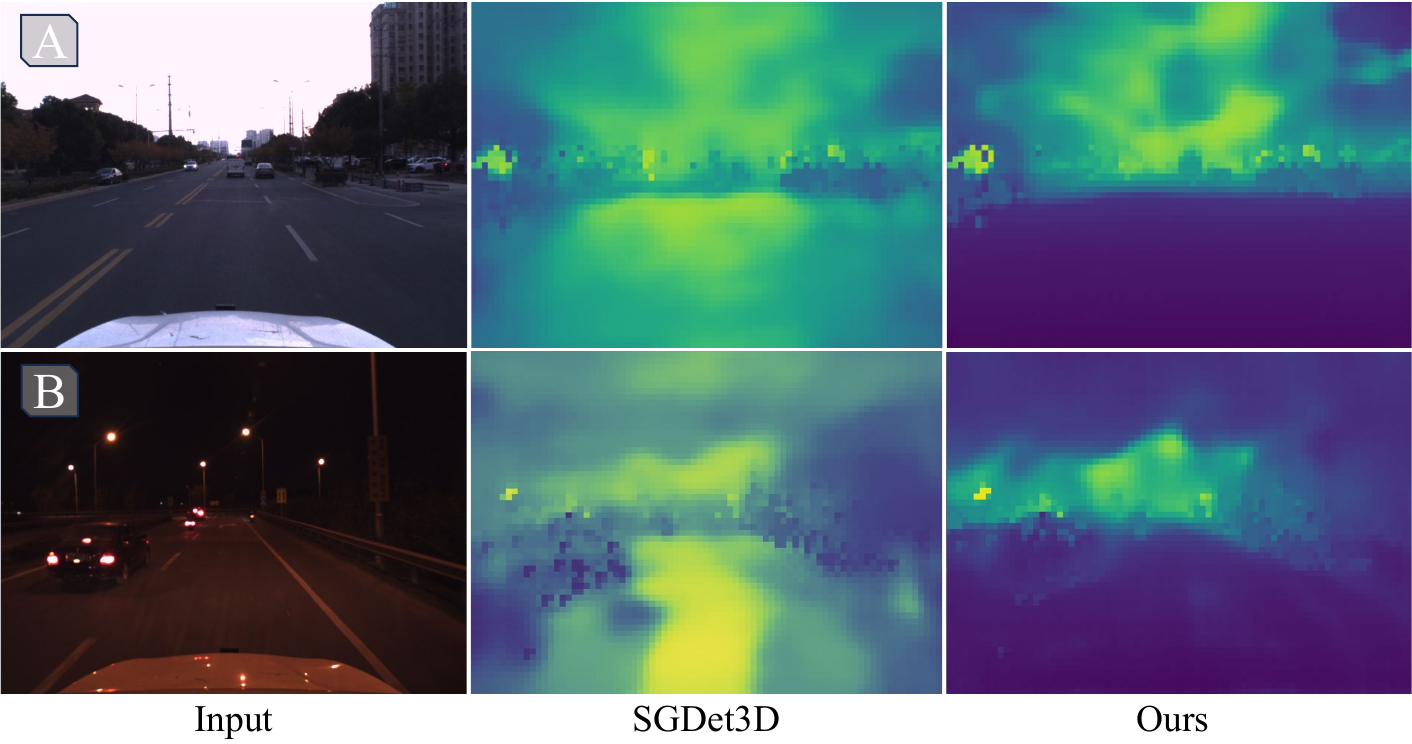}
    \vspace{-5pt}
    \caption{\textbf{Qualitative comparison of depth predictions by our Panoramic Depth Fusion and the baseline method.}  
    }
    \label{fig:tri_vis}
    \vspace{-10pt}
\end{figure}

\noindent\textbf{Effectiveness of the PDF Module.}
\label{sec:tri_depth}
We evaluate the effectiveness of the Panoramic Depth Fusion module under a unified training configuration. 
The module uses four loss weights:
\(\lambda_1\) (0.1),  
\(\lambda_{\mathrm{abs}}\) (0.01),  
\(\lambda_{\mathrm{dense}}\) (0.03), and  
\(\lambda_3\) (0.05).  
Different combinations correspond to different supervision configurations, as summarized in~\tabref{tab:tri_ablation}.  
Qualitative comparisons are provided in~\figref{fig:tri_vis}. Our model maintains accurate depth and clear object boundaries under high-noise and low-light conditions, demonstrating strong generalization.
Adding the metric loss (\(\lambda_{\mathrm{dense}}\)) improves mAP from 45.15 to 46.08 (+0.93), while the relative loss (\(\lambda_3\)) further increases it to 46.86 (+0.78).

\begin{table}[t]
\centering
\caption{\textbf{Ablation of the Panoramic Depth Fusion module.}  
Each setting corresponds to different combinations of loss components. }
\vspace{-10pt}
\label{tab:tri_ablation}
\resizebox{0.9\columnwidth}{!}{
\begin{tabular}{c|cccc|c}
\toprule
\textbf{Setting} & $\lambda_1$ & $\lambda_{\mathrm{abs}}$ & $\lambda_{\mathrm{dense}}$ & $\lambda_3$ & \textbf{BEV mAP}$\uparrow$ \\
\midrule
A  & 0.1 & 0.01 & -- & -- & 45.15 \\
B  & 0.1 & 0.01 & 0.03 & -- & 46.08 \\
C  & 0.1 & 0.01 & 0.03 & 0.05 & \textbf{46.86} \\
\bottomrule
\end{tabular}
}
\vspace{-10pt}
\end{table}

\vspace{2pt}
\noindent\textbf{Effectiveness of the DGTF Module}
\label{sec:dgr_ablation}
As shown in \tabref{tab:dgr_ablation}, we perform a set of step-wise ablation studies to validate the effectiveness of each component in the DGTF module. 
Adding simple feature concatenation (V1) already improves temporal consistency, while deformable alignment (V2) further enhances motion-aware fusion. 
Introducing the gating mechanism (ConvGRU) yields the largest gain (+3.45 3D mAP), confirming the importance of adaptive update and forgetting. 
Squeeze-and-Excitation brings no additional benefit, suggesting that ConvGRU already performs effective channel selection.
Overall, DGTF (V4) enhances BEV localization by +3.55 mAP and 3D detection by +3.45 mAP, demonstrating robust and efficient temporal fusion.

\begin{table}[t]
\centering
\caption{\textbf{Ablation of the Deformable Gated Temporal Fusion module.} 
V1 (+Concat) concatenates current and previous BEV features; 
V2 (+DCN) adds modulated deformable convolution (DCNv2) to spatially align features; 
V4 (DGTF, Ours) further incorporates a GRU-style gating mechanism (ConvGRU) for selective temporal fusion; 
V3 (+SE) adds channel-wise Squeeze-and-Excitation on top of V4.
}
\vspace{-7pt}
\label{tab:dgr_ablation}
\resizebox{0.98\columnwidth}{!}{
\begin{tabular}{l|cccc|cc}
\toprule
\textbf{Variant} & \textbf{Concat} & \textbf{DCN} & \textbf{ConvGRU} & \textbf{SE} & \textbf{BEV mAP} & \textbf{3D mAP} \\
\midrule
V0  & &  &  &  & 46.86 & 41.41 \\
V1  & \checkmark &  &  &  & 47.82 & 42.01 \\
V2   &  \checkmark & \checkmark &  &  & 48.86 & 43.32 \\
V3 & \checkmark & \checkmark   & \checkmark &\checkmark & 49.12 & 42.47 \\
 \textbf{V4 (Ours)}  &  \checkmark & \checkmark & \checkmark & & \textbf{50.41} & \textbf{44.86} \\
\bottomrule
\end{tabular}
}
\vspace{-15pt}
\end{table}



\noindent\textbf{Effectiveness of the IGDR Module}
To validate the effectiveness of each component in the proposed IGDR module, we conduct a comprehensive ablation study. Starting from a baseline that only uses the main BEV feature $F_{RC}$, we progressively build up to the full model.
The results clearly demonstrate the evolution of our IGDR design:
(1) \textit{From baseline to prototype fusion (A$\rightarrow$B$\rightarrow$C):} The baseline A (+PDF+DGTF), which directly uses the raw BEV features $F_{RC}$, achieves 50.41\% mAP. Naively adding the Softmax-weighted prototype (B) yields limited gains, demonstrating that direct fusion introduces background noise. However, introducing the gating mechanism (C) to selectively fuse the foreground significantly improves performance to 51.60\% (+1.19 mAP over A). This confirms that gating is essential to filter out unreliable regions and provides a clean foundation for dynamic refinement.
(2) \textit{Necessity of dynamic calibration (C$\rightarrow$D):} The MLP-based calibration (D) introduces instance-adaptive $\gamma, \beta$ parameters from $E_{proj}$, leading to a further +0.88 mAP gain, validating the idea of calibrating rather than replacing $F_{RC}$.
(3) \textit{Choice of calibration generator (D$\rightarrow$E$\rightarrow$F):} While the attention-based generator (E) provides marginal improvement (+0.04 mAP), our convolutional generator (F) achieves the best result (+1.55 mAP). The spatial-to-spatial mapping ($E_{BEV}$$\rightarrow$F$_{RC}$) via convolution captures local geometric patterns more effectively than non-spatial (MLP) or attention-based counterparts.
Overall, the proposed IGDR achieves the highest performance (54.07 mAP, +3.66 over the baseline) and demonstrates a robust, efficient refinement mechanism for radar-camera BEV fusion.

\begin{table}[t]
\centering
\caption{\textbf{Ablation on the Instance-Guided Dynamic Refinement module.} 
\textit{Gen.} indicates the generator used to produce calibration parameters or replacement maps. 
(1) \textbf{None (A):} Baseline without refinement. 
(2) \textbf{Softmax-Fusion (B):} Directly adds the Softmax-weighted instance prototype $E_{BEV}$ to the main stream without gating. 
(3) \textbf{Static Map (C):} Statically replaces the foreground region in $F_{RC}$ with $E_{BEV}$ under gating. 
(4) \textbf{MLP (D):} Generates $\gamma, \beta$ from $E_{proj}$ using MLPs, followed by affine transformation. 
(5) \textbf{Attention (E):} Employs a masked attention mechanism using $F_{RC}$ as Query and $E_{proj}$ as Key/Value. 
(6) \textbf{Conv (F, Ours):} Constructs the spatial prototype $E_{BEV}$ and generates $\gamma, \beta$ via Conv2d layers, enabling spatially-aware calibration.
Our convolutional generator achieves the best overall performance.}
\label{tab:igdr_ablation}
\vspace{-7pt}
\resizebox{\columnwidth}{!}{
\begin{tabular}{l|cc|ccc}
\toprule
\multirow{2}{*}{\textbf{ID}} & \multicolumn{2}{c|}{\textbf{Component}} & \multicolumn{3}{c}{\textbf{BEV mAP (\%)}} \\
\cmidrule{2-6}
& $G_{bg}$ & Gen. & Cyclist & Pedestrian & Overall \\
\midrule
A &  & None & 57.51 & 31.66 & 50.41 \\
B &  & Softmax-Fusion & 58.14 & 32.00 & 51.29 \\
C & \checkmark & Static Map & 58.62 & 31.92 & 51.60 \\
D & \checkmark & MLP & 62.13 & 32.43 & 52.48 \\
E & \checkmark & Attention & 62.21 & 32.21 & 52.52 \\
F & \checkmark & \textbf{Conv} & \textbf{64.59} & \textbf{33.48} & \textbf{54.07} \\
\bottomrule
\end{tabular}
}
\vspace{-7pt}
\end{table}

\section{Conclusion}
\label{sec:conclusion}
In this paper, we present \ours, a 4D radar-camera multi-modal 3D object detector for highly accurate, efficient, and robust 3D object detection.
\ours~enhances depth estimation quality via the Panoramic Depth Fusion module, enabling mutual reinforcement between absolute and relative depth. 
For temporal fusion, we design a Deformable Gated Temporal Fusion module that does not rely on the ego vehicle's pose.
In addition, we built an Instance-Guided Dynamic Refinement module that extracts semantic prototypes from 2D instance guidance. 
Experiments show that \ours~achieves state-of-the-art 3D object detection results on both TJ4DRadSet and VoD datasets.


\section*{Acknowledgements}

This work was supported by the National Natural Science Foundation of China (Grant No. 62176007).
{
    \small
    \bibliographystyle{ieeenat_fullname}
    \bibliography{main}

\begin{thebibliography}{29}
\providecommand{\natexlab}[1]{#1}
\providecommand{\url}[1]{\texttt{#1}}
\expandafter\ifx\csname urlstyle\endcsname\relax
  \providecommand{\doi}[1]{doi: #1}\else
  \providecommand{\doi}{doi: \begingroup \urlstyle{rm}\Url}\fi

\bibitem[Bai et~al.(2024)Bai, Yu, Zheng, Zhang, Zhou, Zhang, Wang, Bai, and Shen]{sgdet3d}
Xiaokai Bai, Zhu Yu, Lianqing Zheng, Xiaohan Zhang, Zili Zhou, Xue Zhang, Fang Wang, Jie Bai, and Hui-Liang Shen.
\newblock Sgdet3d: Semantics and geometry fusion for 3d object detection using 4d radar and camera.
\newblock \emph{RAL}, 2024.

\bibitem[Bhat et~al.(2023)Bhat, Birkl, Wofk, Wonka, and M{\"u}ller]{bhat2023zoedepth}
Shariq~Farooq Bhat, Reiner Birkl, Diana Wofk, Peter Wonka, and Matthias M{\"u}ller.
\newblock Zoedepth: Zero-shot transfer by combining relative and metric depth.
\newblock \emph{arXiv preprint arXiv:2302.12288}, 2023.

\bibitem[Chen et~al.(2023)Chen, Zhang, Wang, Wang, and Zhao]{futr3d}
Xuanyao Chen, Tianyuan Zhang, Yue Wang, Yilun Wang, and Hang Zhao.
\newblock Futr3d: A unified sensor fusion framework for 3d detection.
\newblock In \emph{CVPR}, 2023.

\bibitem[Han et~al.(2024)Han, Yang, Sun, Ge, Dong, Zhou, Mao, Peng, and Zhang]{VideoBEV}
Chunrui Han, Jinrong Yang, Jianjian Sun, Zheng Ge, Runpei Dong, Hongyu Zhou, Weixin Mao, Yuang Peng, and Xiangyu Zhang.
\newblock Exploring recurrent long-term temporal fusion for multi-view 3d perception.
\newblock In \emph{RAL}, 2024.

\bibitem[Huang et~al.(2021)Huang, Huang, Zhu, Ye, and Du]{BEVDet}
Junjie Huang, Guan Huang, Zheng Zhu, Yun Ye, and Dalong Du.
\newblock Bevdet: High-performance multi-camera 3d object detection in bird-eye-view.
\newblock \emph{arXiv preprint arXiv:2112.11790}, 2021.

\bibitem[Jiang et~al.(2024)Jiang, Li, Liu, Wang, Jia, Wang, Han, and Zhang]{jiang2024far3d}
Xiaohui Jiang, Shuailin Li, Yingfei Liu, Shihao Wang, Fan Jia, Tiancai Wang, Lijin Han, and Xiangyu Zhang.
\newblock Far3d: Expanding the horizon for surround-view 3d object detection.
\newblock In \emph{AAAI}, 2024.

\bibitem[Kim et~al.(2023{\natexlab{a}})Kim, Kim, Choi, and Kum]{CRAFT}
Youngseok Kim, Sanmin Kim, Jun~Won Choi, and Dongsuk Kum.
\newblock Craft: Camera-radar 3d object detection with spatio-contextual fusion transformer.
\newblock In \emph{AAAI}, 2023{\natexlab{a}}.

\bibitem[Kim et~al.(2023{\natexlab{b}})Kim, Shin, Kim, Lee, Choi, and Kum]{CRN}
Youngseok Kim, Juyeb Shin, Sanmin Kim, In-Jae Lee, Jun~Won Choi, and Dongsuk Kum.
\newblock Crn: camera radar net for accurate, robust, efficient 3d perception.
\newblock In \emph{ICCV}, 2023{\natexlab{b}}.

\bibitem[Lang et~al.(2019)Lang, Vora, Caesar, Zhou, Yang, and Beijbom]{pointpillars}
Alex~H Lang, Sourabh Vora, Holger Caesar, Lubing Zhou, Jiong Yang, and Oscar Beijbom.
\newblock Pointpillars: Fast encoders for object detection from point clouds.
\newblock In \emph{CVPR}, 2019.

\bibitem[Lei et~al.(2023)Lei, Chen, Jia, and Zhang]{HVDetFusion}
Kai Lei, Zhan Chen, Shuman Jia, and Xiaoteng Zhang.
\newblock Hvdetfusion: A simple and robust camera-radar fusion framework.
\newblock \emph{arXiv preprint arXiv:2307.11323}, 2023.

\bibitem[Li et~al.(2023)Li, Ge, Yu, Yang, Wang, Shi, Sun, and Li]{BEVDepth}
Yinhao Li, Zheng Ge, Guanyi Yu, Jinrong Yang, Zengran Wang, Yukang Shi, Jianjian Sun, and Zeming Li.
\newblock Bevdepth: Acquisition of reliable depth for multi-view 3d object detection.
\newblock In \emph{AAAI}, 2023.

\bibitem[Li et~al.(2022)Li, Wang, Li, Xie, Sima, Lu, Qiao, and Dai]{BEVFormer}
Zhiqi Li, Wenhai Wang, Hongyang Li, Enze Xie, Chonghao Sima, Tong Lu, Yu Qiao, and Jifeng Dai.
\newblock Bevformer: Learning bird’s-eye-view representation from multi-camera images via spatiotemporal transformers.
\newblock In \emph{ECCV}, 2022.

\bibitem[Liang et~al.(2022)Liang, Xie, Yu, Xia, Lin, Wang, Tang, Wang, and Tang]{bevfusion}
Tingting Liang, Hongwei Xie, Kaicheng Yu, Zhongyu Xia, Zhiwei Lin, Yongtao Wang, Tao Tang, Bing Wang, and Zhi Tang.
\newblock Bevfusion: A simple and robust lidar-camera fusion framework.
\newblock In \emph{NeurIPS}, 2022.

\bibitem[Lin et~al.(2024)Lin, Liu, Xia, Wang, Wang, Qi, Dong, Dong, Zhang, and Zhu]{RCBEVDet}
Zhiwei Lin, Zhe Liu, Zhongyu Xia, Xinhao Wang, Yongtao Wang, Shengxiang Qi, Yang Dong, Nan Dong, Le Zhang, and Ce Zhu.
\newblock Rcbevdet: Radar-camera fusion in bird’s eye view for 3d object detection.
\newblock In \emph{CVPR}, 2024.

\bibitem[Liu et~al.(2022)Liu, Wang, Zhang, and Sun]{PETR}
Yingfei Liu, Tiancai Wang, Xiangyu Zhang, and Jian Sun.
\newblock Petr: Position embedding transformation for multi-view 3d object detection.
\newblock In \emph{ECCV}, 2022.

\bibitem[Long et~al.(2023)Long, Kumar, Morris, Liu, Castro, and Chakravarty]{RADIANT}
Yunfei Long, Abhinav Kumar, Daniel Morris, Xiaoming Liu, Marcos Castro, and Punarjay Chakravarty.
\newblock Radiant: Radar-image association network for 3d object detection.
\newblock In \emph{AAAI}, 2023.

\bibitem[Palffy et~al.(2022)Palffy, Pool, Baratam, Kooij, and Gavrila]{vod}
Andras Palffy, Ewoud Pool, Srimannarayana Baratam, Julian~FP Kooij, and Dariu~M Gavrila.
\newblock Multi-class road user detection with 3+ 1d radar in the view-of-delft dataset.
\newblock \emph{RAL}, 2022.

\bibitem[Philion and Fidler(2020)]{LSS}
Jonah Philion and Sanja Fidler.
\newblock Lift, splat, shoot: Encoding images from arbitrary camera rigs by implicitly unprojecting to 3d.
\newblock In \emph{ECCV}, 2020.

\bibitem[Stone et~al.(2021)Stone, Maurer, Ayvaci, Angelova, and Jonschkowski]{smurf}
Austin Stone, Daniel Maurer, Alper Ayvaci, Anelia Angelova, and Rico Jonschkowski.
\newblock Smurf: Self-teaching multi-frame unsupervised raft with full-image warping.
\newblock In \emph{CVPR}, 2021.

\bibitem[Wang et~al.(2021)Wang, Guizilini, Zhang, Wang, Zhao, and Solomon]{DETR3D}
Yue Wang, Vitor Guizilini, Tianyuan Zhang, Yilun Wang, Hang Zhao, and Justin Solomon.
\newblock Detr3d: 3d object detection from multi-view images via 3d-to-2d queries.
\newblock In \emph{CoRL}, 2021.

\bibitem[Wolters et~al.(2025)Wolters, Gilg, Teepe, Herzog, Laouichi, Hofmann, and Rigoll]{HyDRa}
Philipp Wolters, Johannes Gilg, Torben Teepe, Fabian Herzog, Anouar Laouichi, Martin Hofmann, and Gerhard Rigoll.
\newblock Unleashing hydra: Hybrid fusion, depth consistency and radar for unified 3d perception.
\newblock In \emph{ICRA}, 2025.

\bibitem[Xiong et~al.(2023)Xiong, Liu, Huang, Han, Xia, and Zhu]{lxl}
Weiyi Xiong, Jianan Liu, Tao Huang, Qing-Long Han, Yuxuan Xia, and Bing Zhu.
\newblock Lxl: Lidar excluded lean 3d object detection with 4d imaging radar and camera fusion.
\newblock \emph{IEEE TIV}, 2023.

\bibitem[Yang et~al.(2023)Yang, Chen, Tian, Tao, Zhu, Zhang, Huang, Li, Qiao, Lu, et~al.]{bevformerv2}
Chenyu Yang, Yuntao Chen, Hao Tian, Chenxin Tao, Xizhou Zhu, Zhaoxiang Zhang, Gao Huang, Hongyang Li, Yu Qiao, Lewei Lu, et~al.
\newblock Bevformer v2: Adapting modern image backbones to bird's-eye-view recognition via perspective supervision.
\newblock In \emph{CVPR}, 2023.

\bibitem[Yang et~al.(2024)Yang, Kang, Huang, Xu, Feng, and Zhao]{yang2024depth}
Lihe Yang, Bingyi Kang, Zilong Huang, Xiaogang Xu, Jiashi Feng, and Hengshuang Zhao.
\newblock Depth anything: Unleashing the power of large-scale unlabeled data.
\newblock In \emph{CVPR}, 2024.

\bibitem[Yin et~al.(2021)Yin, Zhou, and Krahenbuhl]{centerpoint}
Tianwei Yin, Xingyi Zhou, and Philipp Krahenbuhl.
\newblock Center-based 3d object detection and tracking.
\newblock In \emph{CVPR}, 2021.

\bibitem[Yin et~al.(2023)Yin, Zhang, Chen, Cai, Yu, Wang, Chen, and Shen]{yin2023metric3d}
Wei Yin, Chi Zhang, Hao Chen, Zhipeng Cai, Gang Yu, Kaixuan Wang, Xiaozhi Chen, and Chunhua Shen.
\newblock Metric3d: Towards zero-shot metric 3d prediction from a single image.
\newblock In \emph{ICCV}, 2023.

\bibitem[Zheng et~al.(2022)Zheng, Ma, Zhu, Tan, Li, Long, Sun, Chen, Zhang, Wan, et~al.]{tj4dradset}
Lianqing Zheng, Zhixiong Ma, Xichan Zhu, Bin Tan, Sen Li, Kai Long, Weiqi Sun, Sihan Chen, Lu Zhang, Mengyue Wan, et~al.
\newblock Tj4dradset: A 4d radar dataset for autonomous driving.
\newblock \emph{IEEE ITSC}, 2022.

\bibitem[Zheng et~al.(2023)Zheng, Li, Tan, Yang, Chen, Huang, Bai, Zhu, and Ma]{rcfusion}
Lianqing Zheng, Sen Li, Bin Tan, Long Yang, Sihan Chen, Libo Huang, Jie Bai, Xichan Zhu, and Zhixiong Ma.
\newblock Rcfusion: Fusing 4-d radar and camera with bird’s-eye view features for 3-d object detection.
\newblock \emph{IEEE TIM}, 2023.

\bibitem[Zhong et~al.(2025)Zhong, Xiang, Xu, Fu, Xu, Wang, Yang, Pu, and Liu]{cvfusion}
Hanzhi Zhong, Zhiyu Xiang, Ruoyu Xu, Jingyun Fu, Peng Xu, Shaohong Wang, Zhihao Yang, Tianyu Pu, and Eryun Liu.
\newblock Cvfusion: Cross-view fusion of 4d radar and camera for 3d object detection.
\newblock In \emph{ICCV}, 2025.

\end{thebibliography}
}

\clearpage
\setcounter{page}{1}
\maketitlesupplementary

\renewcommand\thesection{\Alph{section}} 
\setcounter{section}{0} 

\begin{figure*}[!b] \centering
    \includegraphics[width=\linewidth]{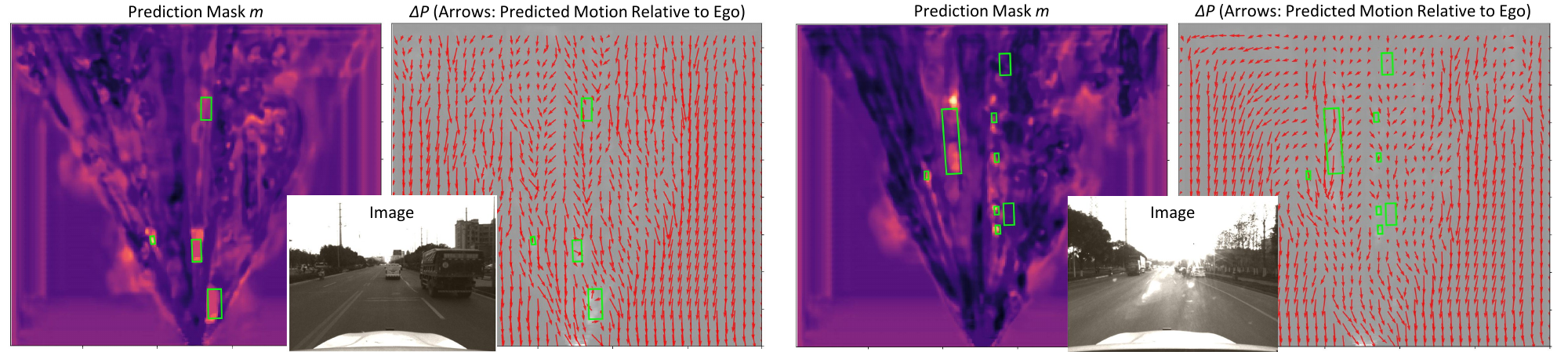}
    \caption{\textbf{DGTF Visualization.} Learned offsets ($\Delta p$, red arrows) align strictly with vehicle motion, while the mask ($m$, heatmap) suppresses background, proving explicit temporal alignment.} 
    \label{fig:dgtf}
\end{figure*}

\section{Additional ablation of the PDF module}
\label{sec:dgtfabl2}

\tabref{tab:tri_ablation2} further presents the experimental process of adjusting the loss weight of the PDF module.
Sensitivity analysis (C1–C4) shows that slight variations of \(\lambda_1\), \(\lambda_3\), \(\lambda_{\mathrm{dense}}\), or \(\lambda_{\mathrm{abs}}\) lead to minor performance drops, confirming that the final setting (C5) achieves a balanced trade-off between stability and accuracy.

\begin{table}[h]
\centering
\caption{\textbf{Ablation of the Panoramic Depth Fusion module.}  
Each setting corresponds to different combinations of loss components. }
\vspace{-10pt}
\label{tab:tri_ablation2}
\resizebox{\columnwidth}{!}{
\begin{tabular}{c|cccc|c}
\toprule
\textbf{Setting} & $\lambda_1$ & $\lambda_{\mathrm{abs}}$ & $\lambda_{\mathrm{dense}}$ & $\lambda_3$ & \textbf{BEV mAP}$\uparrow$ \\
\midrule
A  & 0.1 & 0.01 & -- & -- & 45.15 \\
B (+$\lambda_{\mathrm{dense}}$) & 0.1 & 0.01 & 0.03 & -- & 46.08 \\

C1 (↓$\lambda_1$) & 0.05 & 0.01 & 0.03 & 0.05 & 46.64 \\
C2 (↓$\lambda_3$) & 0.1 & 0.01 & 0.03 & 0.02 & 46.30 \\
C3 (↑$\lambda_\text{dense}$) & 0.1 & 0.01 & 0.05 & 0.05 & 46.45 \\
C4 (↑$\lambda_\text{abs}$) & 0.1 & 0.03 & 0.03 & 0.05 & 46.12 \\
 \textbf{C5 (ours)} & 0.1 & 0.01 & 0.03 & 0.05 & \textbf{46.86} \\
\bottomrule
\end{tabular}
}
\vspace{-10pt}
\end{table}

\section{Additional ablation of the DGTF module}
\label{sec:dgtfabl2}

We also examine temporal depth sensitivity (\tabref{tab:temporal_depth}): using the immediate predecessor ($t-1$) achieves the best accuracy and stability, whereas fusing more distant frames ($t-2$, $t-3$) leads to noise accumulation.

\begin{table}[h]
\centering
\caption{\textbf{Sensitivity to historical frame gap.} Evaluation of different historical fusion depths. Single-step (${t-1}$) achieves the best accuracy and stability. }
\vspace{-7pt}
\label{tab:temporal_depth}
\resizebox{0.75\columnwidth}{!}{
\begin{tabular}{c|cc}
\toprule
\textbf{Fusion Depth} & \textbf{BEV mAP} & \textbf{3D mAP} \\
\midrule
${t~\&~t-3}$ & 49.20 & 42.74 \\
${t~\&~t-2}$ & 49.87 & 43.12 \\
$\boldsymbol{t~ \& ~t-1}$ & \textbf{50.41} & \textbf{44.86} \\
\bottomrule
\end{tabular}
}
\vspace{-7pt}
\end{table}

\section{Qualitative Analyses of the DGTF module}
\label{sec:an2dgtf}

R4Det is the first to validate the feasibility of temporal modeling on these two datasets.
As shown in the~\figref{fig:dgtf}, the learned offsets $\Delta p$ (red arrows) can align with object motion, reconstructing relative motion flow, while the mask $m$ precisely suppresses the background. This physically proves DGTF performs explicit, physics-based motion alignment.

\end{document}